\documentclass[12pt]{gatechthesis}

\title{A Miniature Vision-Based Localization System for Indoor Blimps}
\author{Shicong Ma}
\approvaldate{July 1, 2019}
\school{School of Electrical and Computer Engineering}
\begin{document}

\makeTitlePage{August}{2020}

% In \begin{approvalPage}{N}, the parameter N is the number of members in the committee. If this is less than 4, the layout of the page is single-column rather than two-column, so change the value accordingly.

\begin{approvalPage}{3}

% Add people in the following format:
% \committeeMember{Member Name}{Member Department/Position}{Member Affiliation}

\committeeMember{Dr. Frank Dellaert, Advisor}{School of Interactive Computing}{Georgia Institute of
Technology}
\committeeMember{Dr. Patricio Antonio Vela, Co-Advisor}{School of Electrical and Computer Engineering}{Georgia Institute of Technology}
\committeeMember{Dr. C\'edric Pradalier}{School of Interactive Computing}{Georgia Institute of Technology}
% \committeeMember{Dr. Kimiyo Hoshi}{Division of Science}{Astronomy College of Tokyo}
% \committeeMember{Dr. Michael Holt}{Department of Physics}{Georgia Institute of Technology}
\\ % <- add space for alignment, if necessary
% \committeeMember{Dr. Kent Nelson}{Medical Sciences}{Georgia Institute of Technology}

\end{approvalPage}

\begin{frontmatter}
    
\begin{acknowledgments}
I would like to gratefully acknowledge my supervisor Prof. Frank Dellaert, who provides guidance and feedback throughout my project. Prof. Dellaert gives me the chance to learn much knowledge in both research and programming. \textit{Unit Test Driven Design and Functional Programming}

I would also like to appreciate Prof. C\'edric Pradalier who leads me to the world of probability robotics and all members in the Borg Lab especially Yetong Zhang to provide me with generous help and feedback.

Finally, I would like to acknowledge my dear family for their continuous encouragement and financial support.
\end{acknowledgments}

    \makeTOC
    \makeListOfTables
    \makeListOfFigures
    \newacronym{sfm}{SfM}{Structure from Motion}
\newacronym{vbl}{VBL}{Visual Based Localization}
\newacronym{vo}{VO}{Visual  Odometry}
\newacronym{vslam}{vSLAM}{Visual SLAM}
\newacronym{ate}{ATE}{absolute trajectory error}
\newacronym{dof}{DoF}{Degree of Freedom}
\newacronym{ransac}{RANSAC}{Random Sample Consensus}

\makeListOfAcronyms
    
\begin{summary}

With increasing attention paid to blimp researches, I hope to build an indoor blimp to interact with humans. To begin with, I propose developing a visual localization system to enable blimps to localize themselves in an indoor environment autonomously. This system initially reconstructs an indoor environment by employing Structure from Motion with Superpoint visual features. Next, with the previously built sparse point cloud map, the system generates camera poses by continuously employing pose estimation on matched visual features observed from the map. In this project, the blimp only serves as a reference mobile platform that constraints the weight of the perception system. The perception system contains one monocular camera and a WiFi adaptor to capture and transmit visual data to a ground PC station where the algorithms will be executed. The success of this project will transform remote control indoor blimps into autonomous indoor blimps, which can be utilized for applications such as surveillance, advertisement and indoor mapping.

\end{summary}
\end{frontmatter}

\begin{thesisbody}
    \chapter{Introduction}
The unmanned aerial vehicle (UAV) technology has continuously been evolving since the beginning of the last century with exceptional growth over the previous ten years. Due to UAVs' convenience and high mobility, they have been frequently used for tasks such as aerial crop surveys, aerial photography, search and rescue, and manufacturing/service (e.g., hospitals, greenhouses, production companies, and nuclear power plant).

However, most existing platforms, such as quadcopters, have fast-spinning propellers, which may cause safety concerns in human-occupied indoor environments. Besides, these platforms usually have short flight endurance and generate annoying noises, which poses limitations to their applications. Therefore, a safer robot that can fly for a longer time is increasingly needed.

Blimps, also known as non-rigid airships, can fly significantly longer because they do not require extra energy consumption for buoyancy. Moreover, compared to quad-rotors, blimps have quiet propulsion systems and are less expensive to develop. Therefore, recent researches have shown interest in autonomous blimp development. The main structure of a blimp is a gas (usual helium) balloon and is named as the envelope. The spherical surface makes blimps a safer robot for human interactions. Therefore, blimps have great potential for applications in many fields, such as entertainment, advertising, and search and rescue.

Over the last decade, most research focused on the blimp structure and control system design, while few researchers showed interest in blimp localization. To solve the indoor blimp localization problem with limited payload capacity, \cite{Mueller13phd_Autonomous_Navigation_for_Miniature_Indoor_Airships} and his colleagues developed an accurate sensor fusion localization system. The system localizes the blimp with tiny sonar sensors and microelectromechanical system (MEMS)-based airflow sensors and an inertial measurement unit (IMU). This system had high accuracy, but the cost of building the system was high due to expensive sensors and complicated hardware structures.

Apart from the IMU-based localization system, motivated by camera-based systems used on unmanned aerial robots, there has been a growing interest in applying lightweight and low-cost monocular camera-based localization systems on indoor blimps. But, unlike other unmanned aerial robots (\cite{Kim17icra_astrobee_Robust_Visual_Localization}), especially micro quadcopters (\cite{Lim12cvpr_Real-time_Image-based_6-DOF_Localization_in_Large-Scale_Environments}), only a few works in literature have demonstrated detecting external visual markers with camera (\cite{Fukao03rsj_Autonomous_Blimp_for_Surveillance_System, Yamada09icros_Study_on_SLAM_for_Indoor_Blimp_with_Visual_Markers}) to obtain and show current locations of a robotic blimp. This is due to various difficulties, one of which is the payload constraint. Studies have focused on developing a lightweight camera-based hardware system that can be deployed on a miniature blimp. To achieve efficient interaction between software and hardware, \cite{Al-jarrah13ifac_blimp_embedded_computer_vision_fuzzy_control} presents a complete robotic blimp design that has a camera and communication units mounted to the blimp structure. And to satisfy the payload constraint, the robot captures images with a lightweight camera and transmits data to a ground station where algorithms are executed.

However, the existing localization systems used on indoor blimps rely on external visual markers that are not suitable for large-scale indoor environments. Hence, a system that can efficiently and accurately localize without external markers is in demand. \gls{vbl} has been introduced by researchers to solve robust localization problems with an input video sequence and a prebuilt map. However, the conventional solutions were not efficient and robust during large-scale area exploration and had shown unsatisfying results in low-light environments (\cite{Milford12icra_seqSLAM}). The key issue of the conventional \gls{vbl} system is that the feature descriptors such as SIFT (\cite{Ng03journal_sift}), SURF (\cite{Bay06springer_surf}), and ORB (\cite{Rublee11iccv_orb}) are less robust under lighting changes, hence a more robust descriptor is needed. A new invariant descriptor, SuperPoint, introduced by \cite{Detone18cvpr_SuperPoint}, extracted through the machine learning method, has improved the accuracy of motion blur and huge changes in the field of view feature extraction. This machine learning method can replace the conventional feature extraction method in the \gls{vbl} pipeline to solve the changing illumination vision-based localization problem.

In this paper, a vision-based localization system inspired by \cite{Alcantartilla10icra_learning_visibility}, is developed. Based on \cite{Alcantartilla10icra_learning_visibility}, I project the landmarks in the map to the camera view to filter outliers. And I process the data association process with a searching window around extracted features to match 2D features with 3D landmarks robustly. Furthermore, I formulate a factor graph introduced by \cite{Dellaert17journal_factor_graph_perception} with the data associate information and use GTSAM (\cite{Dellaert12gatech_factor_gtsam}) to solve the pose estimation problem.

In summary, I herein propose solutions for the following problems:
\begin{enumerate}
    \item Assemble a perception system that can attach to a miniature robotic blimp.
    \item Development of an indoor environment vision-based localization solution. 
        \begin{itemize}
            \item Offline sparse map building of an indoor environment.
            \item Continuously 6 \gls{dof} pose estimation with an input video sequence based on a sparse point cloud map and an initial pose prior.
        \end{itemize}
\end{enumerate}
    \chapter{Related Work}

\section{Autonomous Blimp Localization System}

The autonomous blimp is not a new topic, and a large body of literature is available since the late 90s. However, most of the previous studies have been focused on outdoor autonomous blimps. Project “AURORA” in Brazil (\cite{Paiva06journal_aurora}) tried to develop a fully functional outdoor autonomous airship in the late 90s. The visual navigation system developed by “AURORA” uses aerial images as input. After this, several similar projects were launched worldwide, e.g., the Autonomous Airship of LAAS/CNRS (\cite{Lacroix00iace_Toward_autonomous_airships_LAAS, Lacroix02book_The_autonomous_blimp_project_of_LAAS,Hygounenc04sage_autonomous_blimp_project_of_laas}), LOTTE airship in Germany (\cite{Wimmer02ieee_Lotte}), KARI in Korea around a decade ago (\cite{Lee06springer_Korean_high_altitude_platform_systems,Lee04ieee_autonomous_flight_control_system_airship}), and DIVA in Portugal (\cite{Moutinho07journal_Modeling_and_nonlinear_control_for_airship_autonomous_flight}). 

With the progress of technologies such as battery capacity, sensor property, and algorithms, indoor blimps, which are economically friendly and has a wide variety of applications in civil and military fields, have become popular. \cite{Hollinger05thesis_design_indoor_robotic_blimp_for_urban_search} constructed a lighter-than-air airship blimp for use in an urban search and rescue environment. The blimp has motors, batteries, and an undercarriage mounted on the center bottom. And it is attached with a camera, sonar, and side infrareds to follow lines and avoid obstacles. The sensor data are processed in a Linux ground station by communicating through a pair of wireless control modules. This kind of airship like blimp configuration was also adapted by \cite{Gonzalez09rdaf_Developing_low-cost_autonomous_indoor_blimp}. They developed a low-cost autonomous indoor blimp based on a hobby radio-controlled (RC) blimp from Plantraco. But, different from Hollinger, they used ultrasonic sensors to measure the distance from the blimp to obstacles. Later, \cite{Mueller13phd_Autonomous_Navigation_for_Miniature_Indoor_Airships} improved the configuration by redesigning the hull and adding a gondola to carry hardware. He also developed a flexible sensor configuration of an IMU, airflow sensors, and sonar sensors, and replaced the ground station with a lightweight embedded system. \cite{Al-jarrah13ifac_blimp_embedded_computer_vision_fuzzy_control} developed an airship robot that has an embedded system that uses fuzzy logic for obstacle avoidance and a robust embedded visual system to follow a ground robot target by an indoor blimp robot. In a recent study, \cite{Fedorenko16edp_indoor_autonomous_airship_control_and_navigation_system} had described the use of optical flow smart camera and 3-D compass for the position and attitude determination for indoor navigation of an airship.

\section{Vision-Based Localization System}

A vision-based localization system can be divided into global localization and incremental localization. Global localization is to localize a single image with respect to a 3D structure without any prior information. Differently, incremental localization accurately estimates the camera pose over time given an initial estimate.

\gls{vbl} is the method used to solve the global localization problem. For instance, recovering the pose of a camera that took a given photography according to a set of geo-localized images or a 3D model. Furthermore, localizing a robot under a given 3D point cloud, in other words, SLAM loop-closure or relocalization, is also a simple illustration of such a method.

Incremental localization can be further divide into three methods: map based incremental localization system, \gls{vo} (\cite{Zhu08ieee_Real_time_global_localization_with_a_pre_built_visual_landmark_database, Lynen15rss_get_out_my_lab_visual_inertial_loc}), and \gls{vslam} (\cite{Mur-artak16ieee_orbslam2, Engel14eccv_lsd_slam}). \gls{vo} is the process of estimating the ego-motion of an agent(e.g., vehicle, human, and robot) using only a sequence of images. And \gls{vslam} is a process in which a robot localize itself in an unknown environment and build a map of this environment at the same time without any prior information. The combination of different methods is often more robust and efficient than utilizing a single method, but I will only concentrate on each method individually. \gls{vo} is an efficient algorithm but does not solve the drift problem in localization. While vSLAM approaches such as ORB-SLAM (\cite{Mur-artak16ieee_orbslam2}) and LSD-SLAM (\cite{Engel14eccv_lsd_slam}) are increasingly capable, they are not as reliable as techniques which rely on a fixed, pre-computed map. Hence, the flexibility that SLAM provides is unnecessary for fix areas such as indoor environments. Thus if we focus on the accuracy of the retrieval poses, the map-based incremental localization system performs the best in a fixed environment.

\begin{figure}[H]
    \centering
	\includegraphics[width=\textwidth]{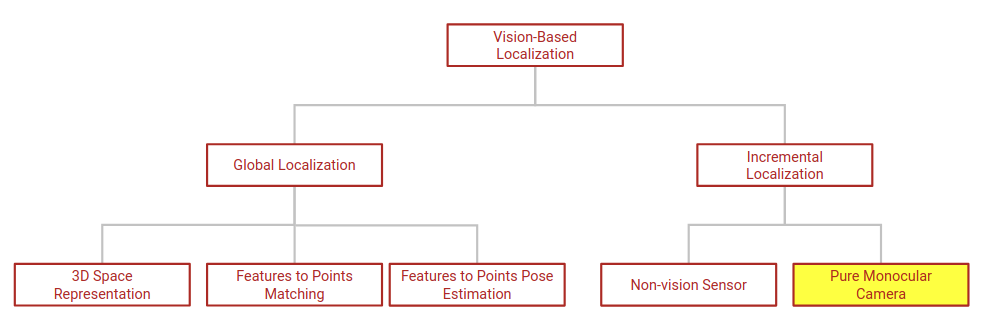}
	\caption[Vision-Based Localization]{Visualization of vision-based localization related work structure}
	\label{fig:vision-based localization}
\end{figure}

As for the following passages, as shown in \ref{fig:vision-based localization}, vision-based localization will be reviewed from two aspects: global localization and incremental localization. As for global localization, I focus on each module of the 3D structure-based VBL pipeline. And as for incremental localization, I concentrate on map-based incremental localization and review from two aspects: systems with non-vision sensors and systems with a pure monocular camera. System with a pure monocular camera, as shown in yellow in \ref{fig:vision-based localization}, is where my research locates.

\subsection{Global Localization}

\gls{vbl} is a broad research topic and has been very active among the computer vision community. Many surveys and reviews have been presented to help researchers to review the existing literature and to narrow down their research interest. \cite{Piasco18pr_survey_vbl_heterogeneous_data} divide VBL into two distinct families: indirect and direct system. Indirect system also known as image retrieval (\cite{Radenovic16eccv_cnn_image_retrieval, Arandjelovic12cvpr_three_image_object_retrieval}) only estimates the coarse camera pose based on a image database. Direct system directly estimate the pose of the query image. Later, \cite{Xin19ricai_review_vbl} categorized and reviewed \gls{vbl} based on three different space representations: image database (\cite{Arandjelovic12cvpr_three_image_object_retrieval,Jegou10cvpr_aggregating,torii15cvpr_24_7_place_recog,Jin15iccv_predicting,Jegou14cvpr_triangulation,cummins08ijrr_fab_map}), learning model (\cite{Shotton13cvpr_scene_regression_forest,Guzman14cvpr_multi_learning_camera_reloc,valentin15cvpr_exploiting_uncertainty_regression_forest,Meng16bmvc_exploiting_rgb_sparse_features,Brachmann16cvpr_uncertainty_driven_6d_pose_estimate}), and 3D structure (\cite{Li12eccv_worldwide_pose_estimation,Zeisl15ieee_CameraPoseVoting,Liu17iccv_efficient,sattler17cvpr_are_large_scale_3d_model_necessary,Irschara09ieee_sfm_Fast_Location_Recognition2,Sattler11iccv_fastLocDirect2D3D}). According to \cite{Xin19ricai_review_vbl}, method based on 3D structure is the most advance VBL method due to their best performance regarding localization accuracy. Hence, we will focus on direct VBL methods that are based on 3D structure.

\subsubsection{3D Space Representation}
Structure-based localization methods assume that a scene is represented by a 3D model. The first step of 3D structure-based VBL is to construct the 3D point cloud model using structure from motion (SfM) algorithm. Open source code such as COLMAP developed by \cite{Schonberger18phdthesis_colmap} and VisualSfM developed by \cite{Wu13ieee_towards} are commonly used to generate SfM reconstruction from database images. Next, each 3D point needs to associate with a descriptor. \cite{Sattler11iccv_fastLocDirect2D3D} evaluated different representation for the 3D points and discovered the best representation could be obtained by using all descriptors or the integer mean of all descriptor per visual word.  

As for the selection of descriptor, several criteria has to be taken into account: scale, orientation and illumination invariance, as well as computational cost and descriptor vector dimension. \cite{Piasco18pr_survey_vbl_heterogeneous_data} provided a comprehensive list of descriptors used in VBL. Based on the list, Hessian-affine detector (\cite{Mikolajczyk04springer_scale_affine}) with SIFT descriptor (\cite{Lowe04springer_sift}) is widely used in image retrieve applications. \cite{Middelberg14ECCV_scalableLocMob} apply RootSIFT (\cite{Arandjelovic12cvpr_three_image_object_retrieval}) to their global localization system to achieve more accurate matching result. \cite{Griffith17wiley_survey} compared outdoor visual data association with BRIEF (\cite{Calonder10eucv_brief}), ORB (\cite{Rublee11ieee_orb}), SIFT, SURF (\cite{Bay08elsevier_surf}). They discovered all these descriptors performed similarly with a slight qualitative advantage for ORB. \cite{Feng15ieee_fast_loc_large_scale_indexing} maintained precision and improved rapidity by using BRISK descriptor (\cite{Leutenegger11ieee_brisk}). 

However, traditional features are not robust to weather and illumination changes. {Krajn\'ik} present a trainable feature GRIEF (Generated BRIEF) to solve image registration under variable lighting and naturally-occurring seasonal changes. \cite{Detone18ieee_SuperPoint} trained SuperPoint from synthetic data and outperform traditional features in feature matching under illumination changes. \cite{Dusmanu19cvpr_d2_net} presented D2-Net and shown significantly better performance under challenging conditions, e.g., when matching daytime and night-time images. \cite{Sarlin19cvpr_robust_hierarchical_localization} based on SuperPoint developed a hierarchical localization system that achieved remarkable localization robustness across large variations of appearance.

\subsubsection{Features to Points Matching}
The next step is to find the correspondences between 2D features and 3D landmark points. The biggest challenge is the efficiency and accuracy. \cite{Irschara09ieee_sfm_Fast_Location_Recognition2} proposed to directly obtain the corresponding content in the 3D model through the feature index of the vocabulary tree, instead of linking the images database. Based on the Irschara's research, \cite{Sattler11iccv_fastLocDirect2D3D} proposed a vocabulary-based priority search (VPS) method inspired by bag of words matching method. In subsequent work, the same authors \cite{Sattler17IEEE_Efficient&EffectiveLargeScaleLoc} increased the VPS framework with points to features matching. \cite{Sattle15iccv_hyperpoints} also introduced a visibility chart to improve positioning accuracy. In order to improve the speed of 2d-3d matching, \cite{Heisterklaus14ieee_image_based_pose_compact_3d_model} introduced MPEG compression method. \cite{Donoser14cvpr_discriminative} trained the random ferns at the top of each point using descriptor redundancy associated with 3D points, and make the speed of 2d-3d matching faster. \cite{Feng15ieee_fast_loc_large_scale_indexing} used a fast point extractor and adopted a binary descriptor in the method, which greatly increases the calculation speed without affecting the accuracy of the pose estimation.

\subsubsection{Features to Points Pose Estimation}
The final step is to estimate the camera pose based on features to points correspondences. Defined by \cite{Hartley03cambridge_multiple}, perspective-n-point (PnP) formulation is the most common tool to recover the absolute camera pose according to the point cloud reconstructed by SfM. \cite{Donoser14cvpr_discriminative}, \cite{Heisterklaus14ieee_image_based_pose_compact_3d_model}, and \cite{ Li10springer_location_prioritized} demonstrated that six correspondences between the image and the 3D model are sufficient to retrieve the pose, if we have no information about the intrinsic parameters of the camera. This formulation is known as P6P and can be solved with Direct Linear Transformation (DLT proposed by \cite{Hartley03cambridge_multiple}). In particular cases, three correspondences between the image and the model are sufficient (P3P pose computation problem). \cite{Irschara09ieee_sfm_Fast_Location_Recognition2} and \cite{Middelberg14ECCV_scalableLocMob} proof the pose estimation problem can be reduced to a P3P formulation if the intrinsic parameters of the camera are known, or if 3 or more \gls{dof} are fixed (\cite{Qu16iap_evaluation_sift_surf, Zeisl15ieee_CameraPoseVoting}). In those particular cases, P3P solver introduced by \cite{Kneip11cvpr_novel_p3p} is mostly used to recover the pose. Works from \cite{Forstner16springer_photogrammetric} and \cite{ Zeisl15ieee_CameraPoseVoting} apply bundle adjustment to refine the initial pose estimation. 

Rather than explicitly estimating a camera pose from 2D-
3D matches, some researchers have proposed learning based methods. \cite{Brubaker13cvpr_lost}, \cite{Brubaker15pami_map}, and \cite{Fernandez13icra_fast} proposed CNN-based approaches directly learn to regress a 6 DoF pose from images. However, as shown by \cite{Fernandez13icra_fast}, such methods do not achieve the same localization accuracy as 3D structure-based algorithms.

\subsection{Incremental localization}
\subsubsection{Incremental Visual Localization with Non-Vision Sensor}

A lot of literature has presented VO systems to estimate camera poses by detecting and tracking 2D features. However, these incremental-based methods only work well for a short time and drift eventually due to accumulate error. To reduce global error, \cite{Zhu08ieee_Real_time_global_localization_with_a_pre_built_visual_landmark_database} utilized IMU and integrate landmark matching to a pre-built landmark database to improve the overall performance of a dual stereo visual odometry system. They used an intelligent subsample method to reduce the size of the database without dropping the accuracy. \cite{Middelberg14ECCV_scalableLocMob} developed a scalable and drift-free image-based localization system which tracked camera locally on a mobile device and align the local map with global one in an external server. In their system, imu is used to exploit gravity information. Based on their research, \cite{Lynen15rss_get_out_my_lab_visual_inertial_loc} presented a large-scale, real-time pose estimation and tracking system that runs on mobile platforms without requiring an external server. Their system employed map and descriptor compression schemes and efficient search algorithm to achieve real-time performance. \cite{Surber17icra_robust_vins_gps} developed a robust localization system by decoupling the local visual-inertial odometry from the global registration to the reference map and utilizing GPS as a weak prior for suggesting loop closures.
 
\subsubsection{Incremental Visual Localization with Pure Monocular Camera}
In contrast to large scale outdoor localization systems that combine camera and non-vision sensors, localization systems with only a monocular camera are commonly used to navigate robots in constraint areas such as the indoor environment. \cite{Alcantartilla10icra_learning_visibility} presented a real-time approach for vision-based localization systems within scenes that have been reconstructed offline using SfM. They explored the visibility of individual landmarks to achieve a much faster algorithm and superior localization results. \cite{Lim12cvpr_Real-time_Image-based_6-DOF_Localization_in_Large-Scale_Environments}  developed a real-time system to continuously compute precise 6-\gls{dof} camera pose, by efficiently tracking natural features and matching them to 3D points in the SfM point cloud. 

Recent methods for indoor visual localization typically aim to achieve robustness to changing lighting conditions by relying on map representations that target illumination invariance. This includes approaches using local feature descriptors, which are invariant to affine changes in illumination. \cite{Kim17icra_astrobee_Robust_Visual_Localization} presents an illumination-robust visual localization algorithm for a free-flying robot designed to navigate on the International Space Station (ISS) autonomously. The online image localization algorithm extracts BRISK (\cite{Leutenegger2011brisk}) descriptors from the query image and match with a prebuilt feature map. Their approach resolved constant lighting changes localization problems but suffered from irregular lighting changes. \cite{Caselitz20icra_Camera_Tracking_in_Lighting_Adaptable_Maps_of_Indoor_Environments} presented a direct dense camera tracking approach and demonstrated its performance in real-world experiments in scenes with varying lighting conditions.
    \chapter{Technical Approach}
In this chapter, the technical approach will be presented to the reader in two sections: mapping and pose estimation. The mapping and pose estimation sections are presented to help the readers understand the underlying theory of my vision-based localization pipeline.

\begin{figure}[H]
    \centering
	\includegraphics[width=\textwidth]{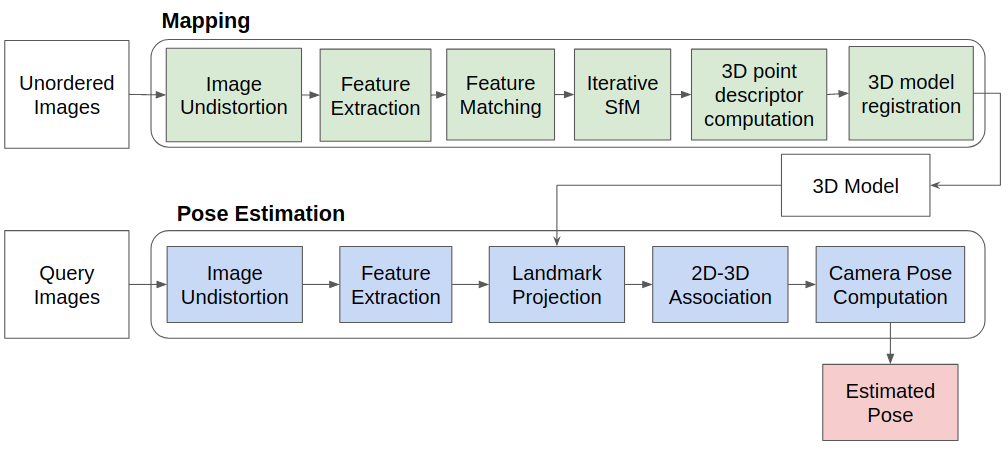}
	\caption[System Pipeline]{This figure is an overview of my localization pipeline.}
	\label{fig:pipeline}
\end{figure}
As shown in \ref{fig:pipeline}, the task is to build a localization pipeline using a 3D structure-based localization method. The pipeline can continuously estimate the camera pose for every image in a given sequence with a 3D model. This pipeline is developed in a modular structure in that each module can be replaced or improved without influencing the remaining modules. 

Before localization, given a set of images, features are extracted from all undistorted images. All features are then matched between every pair of images. The matched features are used to compute the \gls{sfm} reconstruction. Since \gls{sfm} does not generate descriptors for the reconstructed 3D points, a descriptor for each 3D point is calculated by creating an elementwise mean descriptor from all feature descriptors mapped to each 3D point. During localization, 3D landmark points are projected into the estimated pose to efficiently filter 3D points that are not within the field of view. The remaining 3D landmark points are matched with 2D features by computing the descriptor distances. The 2D-3D associations are then used to generate a nonlinear least square problem to estimate the camera pose. 

\section{Mapping}
The goal of mapping is to create a 3D model that best describes a stationary scene. The mapping pipeline contains six main steps:

\begin{enumerate}
\item{Image Undistortion.}
\item{Feature Extraction}
\item{Feature Matching.}
\item{Iterative SfM}
\item{3D Point Descriptor Computation}
\item{3D Model Registration.}
\end{enumerate}

\subsection{Image Undistortion}

Image undistortion is to correct the non-linear projection of the surface points of objects onto the image plane due to lens distortion. There are two common types of distortion, tangential distortion and radial distortion. These two types of distortion can be correct with a five parameters undistortion model.

a. Assuming a point $P=\begin{bmatrix} X & Y & Z \end{bmatrix} ^ T$ in the camera frame, its coordinate in the normalized camera frame is:
\begin{equation}
\begin{bmatrix} x \\ y  \end{bmatrix}=\begin{bmatrix} X/Z \\ Y/Z \end{bmatrix}
\end{equation}

b. Correct tangential distortion and radial distortion in the normalized camera frame:
\begin{equation}
\begin{split}
    x_{distort} = x(1+k_1r^2+k_2r^4+k_3r^6)+2p_1xy+p_2(r^2+2x^2)\\
    y_{distort} = y(1+k_1r^2+k_2r^4+k_3r^6)+p_1(r^2+2y^2)+2p_2xy
\end{split}
\label{eq:3.3.2}
\end{equation}

c. Finally, project the new point into the pixel coordinate will get the correct position of the point in the image.

\subsection{Feature Extraction}

The process of extracting key points and computing descriptors is called feature extraction. Key points are representative points that can describe an image in a compression way. And key points can still be found in an image when the camera field of view sightly changes. Descriptors are information encoded into features to discriminate features from each other. Different features compute key points and descriptors in different ways. The most common feature descriptor is SIFT. Sift key point is obtained by sliding a Difference of Gaussian filter over the image and detecting local extrema for the filter. The descriptor of Sift is a 128 length vector describing the gradients around the key point.

\subsection{Feature Matching}

Feature Matching is to compare feature descriptors across the images to identify similar features. Comparing descriptors is done by computing the Euclidean distance between them. Because descriptors are usually high dimensional vectors, comparing descriptors from two sets of features is expensive. To decrease the matching time, approximate search is used instead of exhaustively looking for the best match. Approximate search is to use a k-dimensional tree (k-d tree) to store all data. In this way, given a descriptor, the best match or the most similar data can be found by following the path from the top of the tree. 

As quoted from \cite{garsten20206dof}:
\begin{quote}
To further ensure that good matches are found it is common to use Lowe's ratio test to discard matches that are probably to be false. The ratio test takes the ratio of the distance between the first and second nearest neighbors of the descriptor to be matched and evaluates if the ratio is less than a pre-defined threshold. The first nearest neighbor is the most similar descriptor to the matched descriptor, while the second nearest neighbor is the second most similar descriptor. This is expressed as
\begin{equation}
    \frac{|d_i - d_{N1}|^2}{|d_i - d_{N2}|^2}\leq tol
\end{equation}
where $d_i$ is the descriptor to be matched, $d_{N1}$ and $d_{N2}$ are the first and second nearest neighbor matches and $tol$ is the predefined tolerance. A match passes the ratio test if the ratio is below the threshold. Employing the ratio test ensures that matched descriptors are more similar to each other than to other descriptors. This filters out matches that are unlikely to be correct. 
\end{quote}

However, in most of the case, a ratio test cannot eliminate false positives. Next, using the RANSCAC algorithm to find the geometry constraint, such as homography matrix, fundamental matrix or essential matrix if the camera calibration matrix is known, can filter bad matches further.

\subsubsection{RANSAC}
\gls{ransac} is a method for picking random data samples to ensures an outlier free subset of data to a certain degree. In many optimization tasks, the data is not perfect. It contains noisy measurements and outliers. Optimizing for all data can then be challenging. Outliers can be far from a correct measurement, contributing to the cost function's large errors, drowning the effect of errors from accurate measurements. The \gls{ransac} method is widely used to try and find a subset of data that only contains inliers.
\begin{equation}
N = \frac{log(1-p)}{log(1-(1-e)^s)}
\label{eq:6}
\end{equation}

\gls{ransac} is solving a problem of using large input data and a solver to compute a solution. The merit of \gls{ransac} is that it can pick a subset of data within the large dataset, which is nearly outlier free, resulting the correct solution's computation. \gls{ransac} randomly picks a subset of $s$ data points $N$ number of times. $N$ is calculated by \ref{eq:6}, where  $e$ is the ratio between outliers and all matched points, and $p$ represents the desired probability of finding the best solution or model. The outlier ratio $e$ is hard to determine and scarcely available. It can, therefore, be necessary to update the distribution based on inlier results calculated during runtime. 

For each iteration, it calculates a solution based on the $s$ data points. Then it calculates an error value for each point within the $s$ points by fitting the point into the solution. For points with an error smaller than a pre-determined threshold are considered as inliers. This procedure is done for all points and will give the number of inliers for a given solution. The number of inliers is used to calculate a new outlier ratio and is then inserted back in \ref{eq:6} to calculate the number of iterations needed. Now, each subset $s$ will be associated with a number of inliers and a solution. After all the iterations, the subset $s$ with the maximum inliers is considered as the best subset and the solution associated with the subset is considered the final solution.

\subsection{Iterative SfM}

There are two ways to reconstruct the 3D points and compute the camera poses, incrementally or globally. Global SfM estimates all camera positions and 3D points at the same time. In contrast, incremental SfM adds on one image at a time to grow the reconstruction. Incremental SfM first generate an initial model with a pair of images. New correspondences are found by matching descriptors of new images with the initial model image database. Then 3D points triangulated by new correspondences are added to this initial model. As more images are added, small errors will accumulate, distorting the model. It is common to perform optimization on both 3D point and camera poses with set intervals.   

As quoted from \cite{garsten20206dof}:
\begin{quote}
Optimization or more specifically, bundle adjustment is performed to minimize growing errors by shifting the position of 3D points and, depending on how much information about the camera is known, tuning the projection matrix. Localization error is usually quantified as the reprojection error of 3D points. It is computed by taking the square of all 2D image points subtracted by their respectively reprojected 3D point correspondence as

\begin{equation}
    e = \sum_{i=1}^{n}\sum_{j=1}^{m}\norm{(x_{ij} - \frac{p^1_i\mathbf{X}_j}{p^3_i\mathbf{X}_j}),(y_{ij} -\frac{p^2_i\mathbf{X}_j}{p^3_i\mathbf{X}_j})}^2
\end{equation}
\end{quote}

\subsection{3D Point Descriptor Computation}
The resulting data from the reconstructions provide 3D points in the SfM coordinate. These 3D points are only described in space $(X,Y,Z)$ and do not have a descriptor assigned to them. Hence the 3D descriptor $\bar{d}$ is computed by taking the mean of the feature point descriptors $d_i$ that have been matched to the 3D point. The resulting $\bar{d} = \frac{1}{n}\sum_{i=0}^{n}d_i $ is assigned to the 3D point. The mean is used because it is one of the best representations for 3D points, according to \cite{Sattler11iccv_fastLocDirect2D3D}. This results in a 256-long vector describing the 3D-point.

\subsection{3D Model Registration}

The 3D model registration is to constraint the seven degrees of freedom of the 3D model. These seven degrees of freedom includes three degrees of freedom in transition, three degrees of freedom in orientation, and one degree of freedom in scale. A similarity transform will be applied to rescale the model and transform it into a new position and orientation. 

\begin{equation}
    T_s = \begin{bmatrix} sR & t \\ 0 & 1\end{bmatrix}
\end{equation}
where $s$ is the scaling factor, $R$ is a 3x3 rotation matrix and $t$ is a 3x1 translation vector. 

\section{Pose estimation}
The methodology behind this is to find the current frame observed landmark points in the sparse map and estimate the pose by employing a nonlinear solver on the current pose. The localization system can be described within the Bayesian filtering framework. 
$$\underbrace{P(X_t|Z^{1:t},L)}_{Posterior} \propto \underbrace{P(z_t|X_t, L)}_{Measurement} \cdot \int_{\theta_{t-1}} \underbrace{P(X_{t}|X_{t-1})}_{Motion} \underbrace{P(X_{t-1}|Z^{t-1},L)}_{Prior}dX_{t-1} $$
Where L is a set of high quality landmarks reconstructed from the images and $Z^{1:t} \equiv z_t$ indicates the sequence of images up to time t.

There are 5 modules in pose estimation:

\begin{enumerate}
\item{Image Undistortion}
\item{Feature Extraction}
\item{Landmark Projection}
\item{2D-3D Association}
\item{Camera Pose Computation}
\end{enumerate}

The first two are the same as in the mapping section, so we only cover the last three below:

\subsection{Landmark Projection}

\begin{figure}[H]
    \centering
	\includegraphics[width=0.7\textwidth]{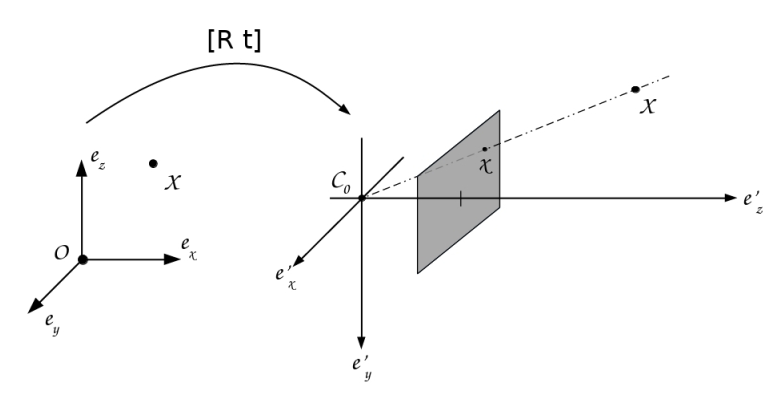}
	\caption[Camera Matrix]{Visualization of applying the camera extrinsics will transform a 3D point from global coordinates into the camera coordinate system. Image adopt from \cite{garsten20206dof}}
	\label{fig:camera matrix}
\end{figure}

\begin{figure}[H]
    \centering
	\includegraphics[width=0.7\textwidth]{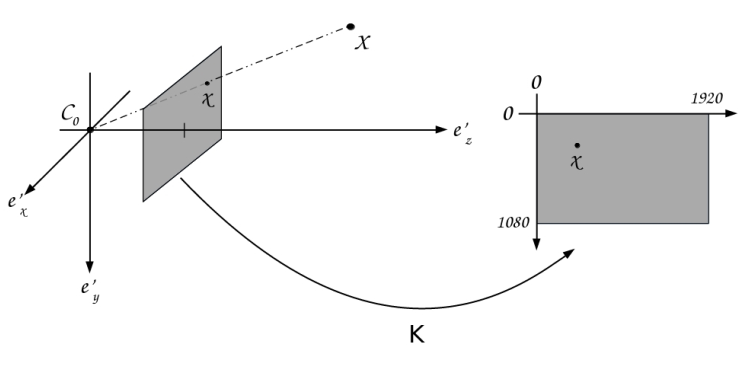}
	\caption[Calibration Matrix]{Applying the calibration matrix K to a point in camera coordinates will reproject it into the image as a pixel coordinate. Image adopt from \cite{garsten20206dof}}
	\label{fig:calibration matrix}
\end{figure}

The camera model describes the process of projecting a point in the 3D world to a 2D image plane. There are many types of camera models. Different camera models are used to describe different kinds of camera lenses. The simplest camera model is the pinhole camera model. The pinhole camera model is a very common and efficient model that describes image rays pass through a pinhole from the front of the pinhole and generates an image by intersecting the image plane at the back of the pinhole. The pinhole is the camera center where all image rays intersect. 

An image pixel coordinate $\begin{bmatrix} u & v \end{bmatrix}^T$ can be calculated by projecting the global point $\begin{bmatrix}X & Y & Z\end{bmatrix}^T$ into the image with a projection matrix $P$:
\begin{equation}
\lambda \underbrace{\begin{bmatrix} u \\ v \\ 1\end{bmatrix}}_{\bar{x}} = P \underbrace{\begin{bmatrix} X \\ Y \\ Z \\ 1\end{bmatrix}}_{\bar{X}}
\label{eq:3}
\end{equation}
where $\lambda$ is a scaling parameter. $\begin{bmatrix} u & v & 1\end{bmatrix}^T$ and $\begin{bmatrix}X & Y & Z & 1\end{bmatrix}^T$ are represented in homogeneous coordinates. Homogeneous coordinates lift an N-dimensional point to an N+1 dimensional line. The N-dimensional point can be retrieved after a transformation by dividing by the last entry and then dropping the last entry:
\begin{equation}
P = K \begin{bmatrix}R & t\end{bmatrix}
\label{eq:1}
\end{equation}
where R is a 3x3 rotation matrix, t is a 3x1 translation vector, and K is a 3x3 calibration matrix. This compact way of applying $R$, $t$ and $K$ requires the use of homogeneous coordinates. $R$ and $t$ are called the camera extrinsics, which provides the camera orientation and position information. $K$ is the camera intrinsics, which describes the property of the camera. Unlike the camera extrinsics, the camera intrinsics do not change when the camera is moving. As seen in \ref{fig:camera matrix}, the rotation $R$ and translation $t$ transform a point from a world coordinate system into a camera coordinate system. As seen in \ref{fig:calibration matrix}, from a camera coordinate system, points can then be re-projected into the pixel coordinate system by applying the camera calibration matrix $K$ because the calibration matrix transformed distances into pixel values. The calibration matrix for a pinhole camera looks like
\begin{equation}
K = \begin{bmatrix} f_x & s & c_x \\ 0 & f_y & c_y \\ 0 & 0 & 1\end{bmatrix}
\label{eq:2}
\end{equation}
where the parameters are; $f_x$ and $f_y$ are the focal lengths along the x and y coordinate axes of the pixel coordinate system. Skew $s$, which describes the tilt of the pixels in the image. Finally, $c_x$ and $c_y$ maps the origin from where the z-axis intersects the image plane to the upper left pixel.

\subsection{2D-3D Association}

The 2D-3D association is to match 2D features with 3D landmarks. 2D features can be matched with 3D landmarks by comparing the L2 distance of their descriptors or comparing the pixel distance of 3D landmark projected features with the extracted features. Same as the feature matching process, before the comparison of descriptors, data are arranged into a k-d tree. The best match is the descriptor pair with the smallest L2 distance. However, this method is highly inaccurate, especially when a lot of 3D landmark points are similar in descriptors. To efficiently filter the outliers, the geometry relationship between 3D landmarks and 2D features can be taken into account. Since the projected point of the matched 3D landmark point should be close to the extracted feature point. Projected points that are far away from their corresponding extracted feature points can be rejected.

\subsection{Camera Pose Computation}

\begin{figure}[H]
\centering
\includegraphics[width=0.7\textwidth]{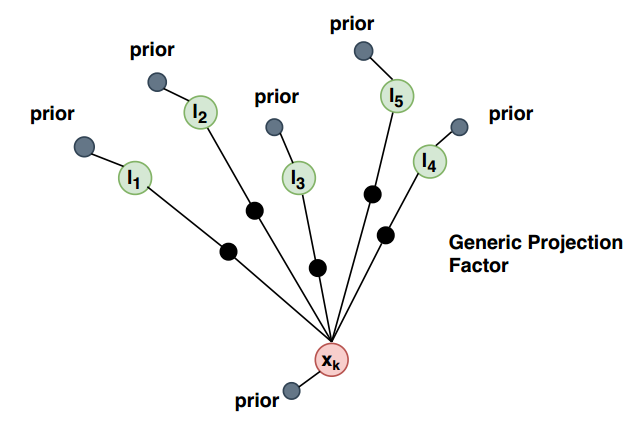}
\caption[Pose Estimation Factor Graph]{Pose Estimation Factor Graph. The Factor Graph contains five prior factors, five landmarks \(\ell_i\), and one camera pose \(x_k\). The estimation of \(x_k\) is the pose of the previous state. The generic projection factors represented by the black nodes are the 2D to 3D correspondences.}
\label{fig:pose estimation}
\end{figure}

Camera pose can be computed by solving a nonlinear least square problem formulated by the 2D-3D association. As shown in \ref{fig:pose estimation}, the nonlinear least square problem can be presented as a factor graph. Each 2D-3D correspondence formulates a generic projection factor. Each point variable is attached to a prior factor.

The error function of a projection factor: 
$$ e_{projection} = ||K(RX+t) - p||_2$$
where $K$ is the camera calibration matrix, $R$ and $t$ is the camera rotation matrix and translation matrix. $p$ is the extracted feature point.

The error function of a prior factor: 
$$ e_{prior} = ||\hat{P} - P||_{\sum}$$
where \^{P} and P can be both 3 dimension point or 6 dimension pose.

The optimization Equation:
$$ X^{MAP} = \underset{X}{argmin} \sum_i (e_{i})^2 $$

\subsubsection{Factor Graph}
As quoted from \cite{Dellaert2012factor}:
\begin{quote}
Factor graphs are graphical models (\cite{koller2009probabilistic}) that are well suited to modeling complex estimation problems, such as Simultaneous Localization and Mapping or \gls{sfm}. You might be familiar with another often used graphical model, Bayes networks, which are directed acyclic graphs. A factor graph, however, is a bipartite graph consisting of factors connected to variables. The variables represent the unknown random variables in the estimation problem, whereas the factors represent probabilistic information on those variables, derived from measurements or prior knowledge.
\end{quote}

    \chapter{Implementation Details}

\section{Blimp Configuration}
The blimp platform is based on an infrared remote control commercial toy blimp, which used a motor control fishtail to provide a forward thrust and a trackpad to adjust its altitude. A AAA battery powers the motor control fishtail and trackpad. Moreover, three additional fins attached on the blimp surface are used to stabilize the blimp while flying.
 
\begin{figure}[H]
    \centering
	\includegraphics[width=\textwidth]{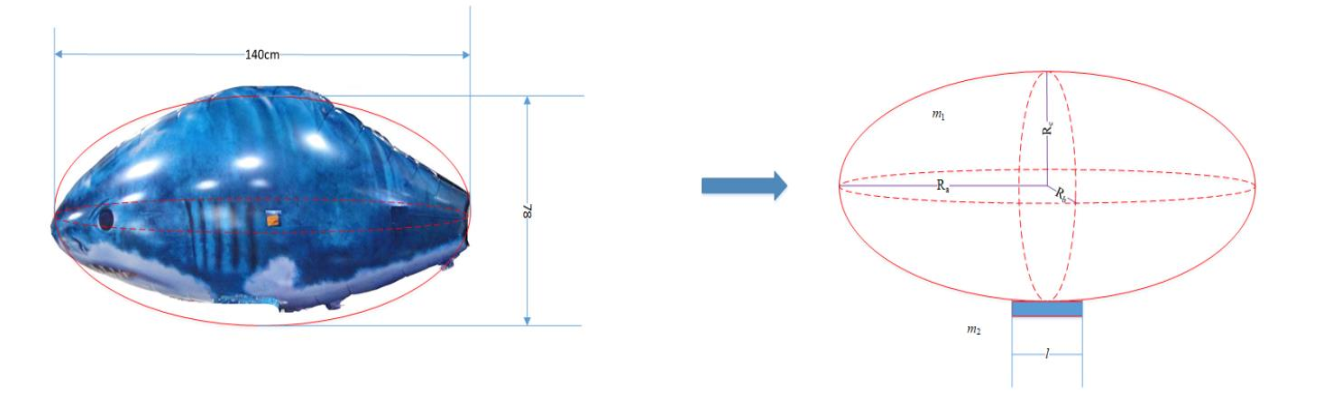}
	\caption[Autonomous Blimp Envelope]{Autonomous Blimp Envelope Figure, adopt from~\cite{Wan18aiaa_design_control_solar_power_blimp}}
	\label{fig:Blimp Envelope}
\end{figure}

As it’s shown in Figure 3.1, a flying toy ‘shark’, which is an inflatable balloon, was selected as the envelope. According to \cite{Wan18aiaa_design_control_solar_power_blimp}, the shape of the blimp is approximated by an ellipsoid with $[R_a; R_b; R_c]$ defined as the three semi-major axes. The volume of the envelope is estimated as Vb =\(190.5 cm^3\). According to Archimedes Principle, the maximum payload capacity is calculated by:
\[ m_{load} = V_b\cdot(\rho_{air}-\rho_{helium}) = 202.1g \]
where \(\rho_{helium} = 169.3g/cm^3\) and \(\rho_{air} = 1226.0g/cm^3\).

\begin{figure}[H]
    \centering
	\includegraphics[width=\textwidth]{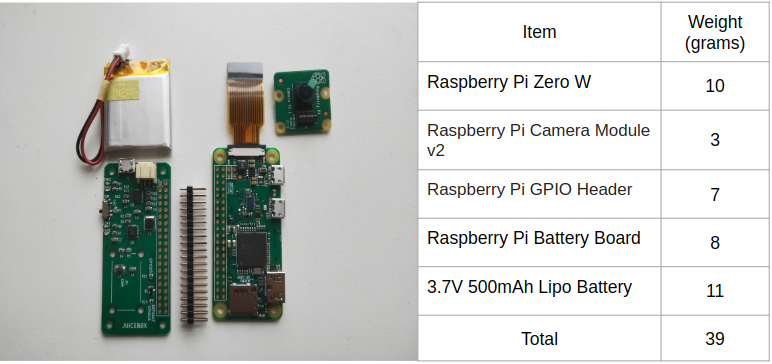}
	\caption[Perception Hardware System]{Perception Hardware System}
	\label{fig:Perception Hardware System}
\end{figure}

I designed a lightweight perception hardware system that includes a raspberry pi zero, a raspberry pi camera, and a Lipo battery, which can be attached to the surface of the envelope.

\begin{figure}[H]
    \centering
	\includegraphics[width=\textwidth]{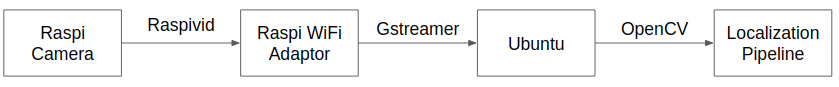}
	\caption[Video Transmission System]{Video Transmission Pipeline}
	\label{fig:Video Transmission Pipeline}
\end{figure}

Video sequences captured by the raspberry pi camera are first processed by Raspivid, which is a package provided by the raspberry pi to access the video captured with the camera module. The video is then transferred to an Ubuntu system in the ground station through WiFi. This process is done with Gstreamer, which is an open-source tool to handle video streaming. After setting up the Gstreamer server and client to establish the connection, OpenCV can be used to capture video streams from the Linux client. 

\section{Mapping}

To reconstruct the 3D model, I first held the perception hardware system (\ref{fig:Perception Hardware System}) by hand and collected unordered images from the Atrium. With the unordered images, I undistorted the images based on the camera calibration matrix through OpenCV. Next, I extracted SuperPoint features and 256-dimensional descriptors that describe the respective points from the undistorted images with the SuperPoint pretrained network. The data is then processed in feature matching by matching the descriptors between every image pairs through OpenCV. I used ratio test and RANSAC introduced by \cite{Fischler1981acm_ransac} to filter bad matches. As for RANSAC, I used a python library called pydegensac which includes LO-RANSAC designed by \cite{Chum03jprs_lo_ransac} and DEGENSAC designed by \cite{Chum05cvpr_two_view_geometry_unaffected_dominant_plane}. This library is marginally better than OpenCV RANSAC. The feature extraction and feature matching data are then stored in a database file.

\begin{figure}[H]
    \centering
	\includegraphics[width=\textwidth]{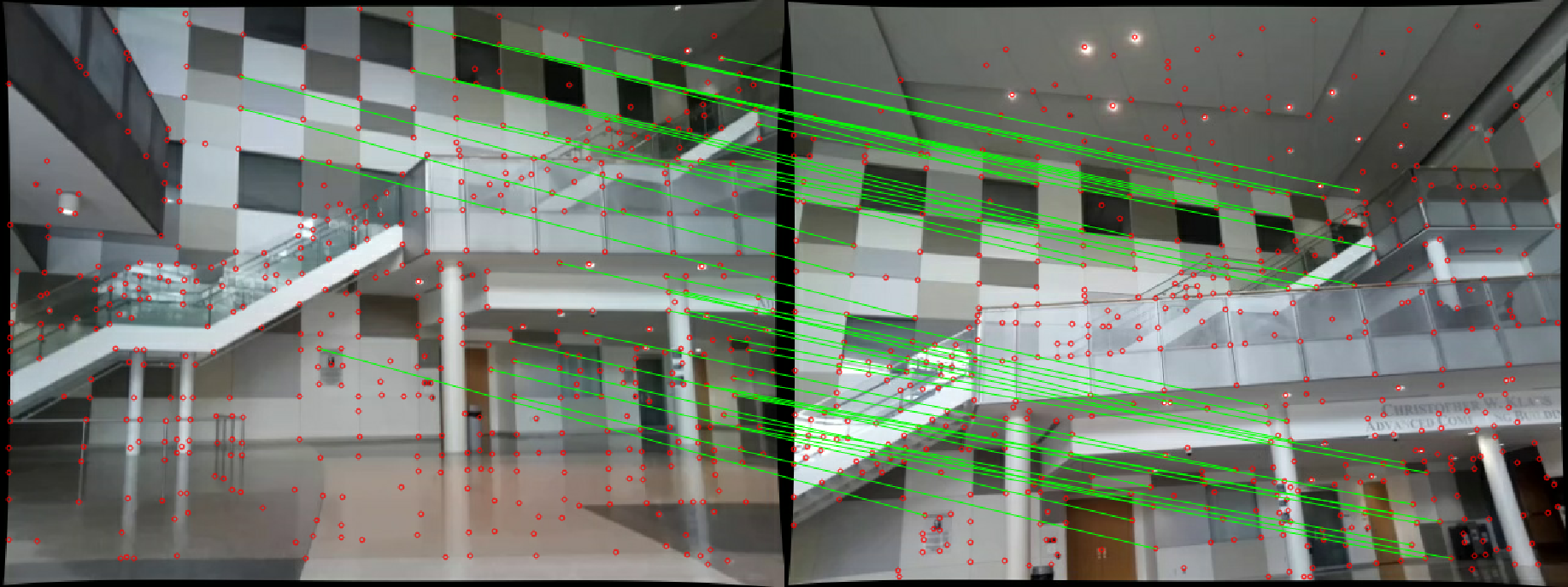}
	\caption[Feature matching]{Feature matching with SuperPoint and pydegensac.}
	\label{fig:feature matching}
\end{figure}

I then stored all data in a database file and used COLMAP developed by \cite{Schonberger18phdthesis_colmap} to finish the final reconstruction through an iterative method. This results in a SuperPoint-based 3D reconstruction. In contrast, to reconstruct the 3D model with Sift features, I do not need to create a database file. I only need to input the unordered images into COLMAP and run COLMAP feature extraction, feature matching, and reconstruction. This is because COLMAP is implemented with Sift feature extraction and matching. The resulting reconstruction, as shown in \ref{fig:Reconstruction Result}, is defined up to an arbitrary scaling factor compared to the real-world 3D model. 

\begin{figure}[H]
    \centering
	\includegraphics[width=\textwidth]{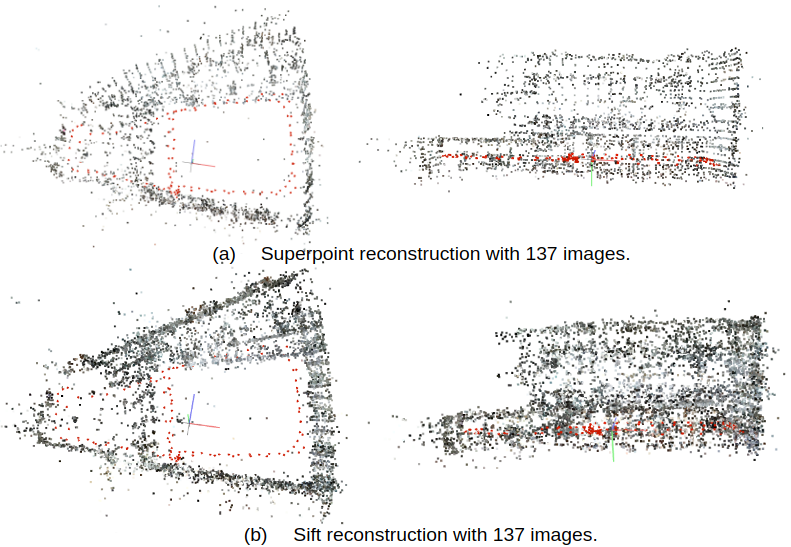}
	\caption[Reconstruction Result]{COLMAP 3D Reconstruction Result with SuperPoint and Sift Descriptors}
	\label{fig:Reconstruction Result}
\end{figure}

The result from COLMAP does not contain the descriptors for the 3D points. Hence, I use the database file and the COLMAP reconstruction output files to create a new data structure named map containing the 3D points and the 3D point descriptors. 

The reconstruction results were then registered to the real world scale by comparing it with the building floorplans. I first selected pairs of points in the reconstructed 3D models and found their correspondences in the building floorplans. Next, I recovered the scale by calculating the point distance differences. The scale is up to a scale factor of 1.2.

\section{Pose Estimation}

\begin{figure}[H]
\centering
\includegraphics[width=0.7\textwidth]{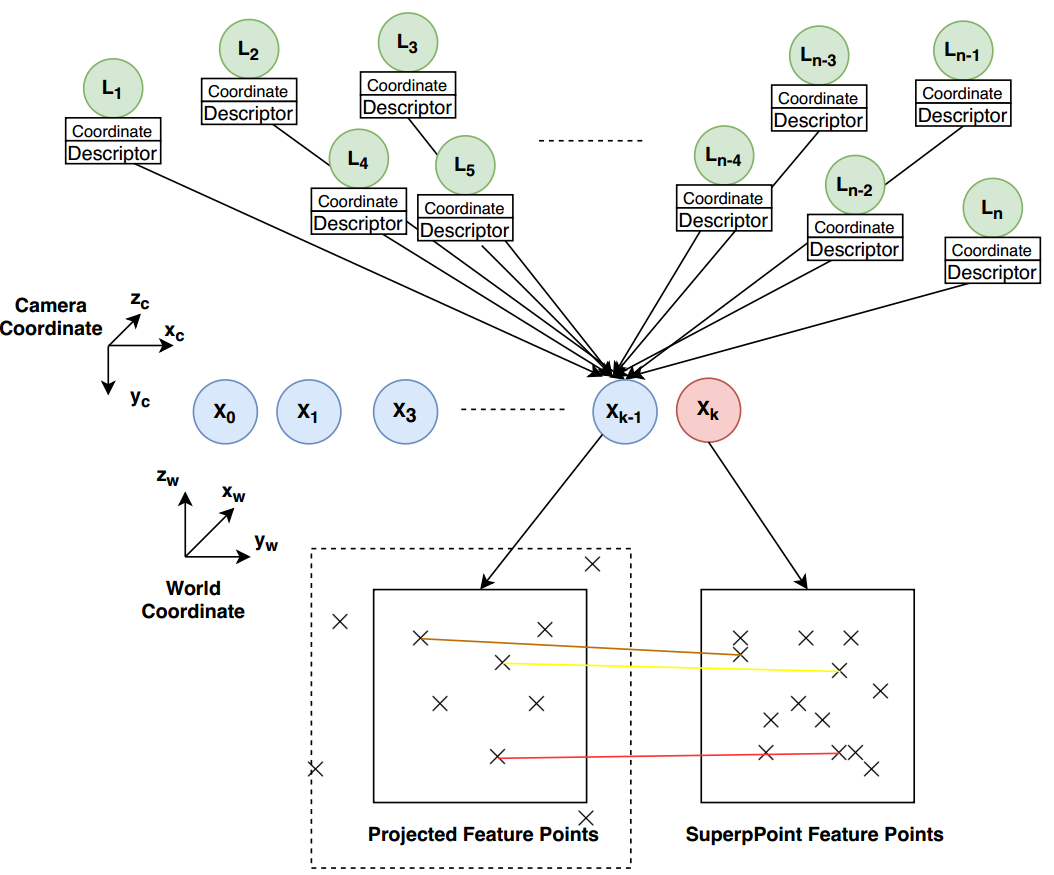}
\caption[Visual-Based Localization]{Camera Tracking. Feature points projected outside the image are invalid projected points.}
\label{fig:vbl}
\end{figure}

The camera tracking or incremental pose estimation process is presented in \ref{fig:vbl}. 3D points from the sparse point cloud are projected to the estimated camera pose (\ref{eq:3}) with points behind the camera or out of the range of the image filtered. In my pipeline, I assumed the camera motion is static due to the slow-motion of a blimp. Hence the 3D points are projected to the pose of the previous state.

\begin{figure}[H]
    \centering
	\includegraphics[width=0.5\textwidth]{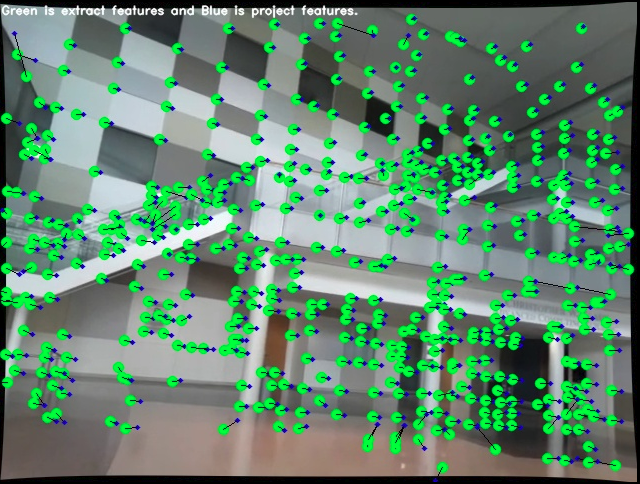}
	\caption[Extract features match with project features]{The blue dots are the projected features, and green circles are the extracted features. The extracted features are displayed in a large green circle for better visualization purposes.}
	\label{fig:Extract feature match with project feature}
\end{figure}

Next, SuperPoint features are extracted from the current undistorted input frame. Measurements, which are the correspondences of 3D landmark points and 2D features, are then obtained by matching between the projected 2D features and the extracted 2D features. The matching criteria are the pixel distance and the L2 distance between descriptors. 

Finally, the current pose is obtained with a nonlinear solver represented by a factor graph, as shown in \ref{fig:pose estimation}. 

\begin{algorithm}[ht]
\SetAlgoLined
\KwResult{An estimate trajectory}
 Read 3D model and initial pose\;
 $trajectory = [initial\_pose]$\;
 \While{True}{
  Read a new frame\;
   Calculate estimate pose based on the camera motion model\;
  \eIf{$frame\:entropy > minimal\:entropy$}{
    Undistort image and extract SuperPoint features\;
    Project 3D landmarks into the estimate pose\;
    Filter projected points that are behind the camera or out of the range of the image\;
    \For{SuperPoint feature in SuperPoint features}{
     Use KNN to find all project landmarks around the SuperPoint feature within a radius threshold\;
     Find the project landmark with the smallest descriptor distance\;
     \eIf{$smallest\:descriptor\:distance < maximum\:distance$}{
        Append 2D feature point and 3D landmark point into an observation list\;
     }
     {
      continue
     }
    }
    \eIf{$number\:of\:observations > minimal\:observation$}{
        Use the observation list to construct a Factor Graph with generic projection factor and Huber noise model\;
        Optimize Factor Graph with Gauss Newton optimizer\;
    }{
    continue\;
    }
  }{
   continue;
  }
 }

 \caption{Visual Based Localization}
\end{algorithm}
    \chapter{Experiments and Results}
In this chapter, evaluation experiments will be performed with the localization system to show the readers how the localization result is affected by different factors, such as illumination, descriptor, 3D model density and, etc. I will first introduce the experiment setup, trajectory error metrics, and ground truth to help readers understand the experiments' process and evaluation metrics.

\section{Experimental Setup}
For all programs and tests, I ran on a standard laptop PC on a 64-bit Ubuntu 18.04 system. The laptop is equipped with an Intel Core i7-9750H 2.60Hzx12 CPU, 16GB DDR, and an Nvidia Geforce RTX 2060 GPU card. The sensor I used is a raspberry pi camera v2.1 (rolling shutter) with a 640x480 resolution and a frame rate of up to 30Hz. All experiments are conducted within the Georgia Institute of Technology Klaus Atrium.

\section{Trajectory Error Metrics}
\cite{Zhang18iros_tutorial_quantitative_evaluation} proposed the evaluation method used for the following experiments. As shown in \ref{fig:ate}, the estimation $\hat{\mathbf{X}}$ is transformed to the aligned estimation $\hat{\mathbf{X^\prime}}$ before the evaluation. Because I am using a monocular camera, this transformation from the estimate trajectory to the aligned estimate trajectory is a similarity transform. Next I calculated the \gls{ate} with the groundtruth $\mathbf{X}_{gt}$ and the aligned estimation $\hat{\mathbf{X^\prime}}$.

\begin{figure}[ht]
    \centering
	\includegraphics[width=0.7\textwidth]{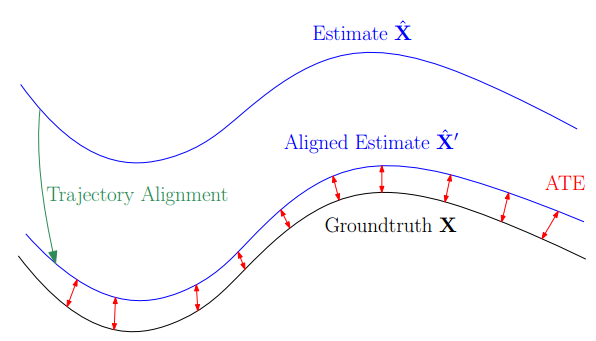}
	\caption[Absolute Trajectory Error]{ATE, adopt from~\cite{Zhang18iros_tutorial_quantitative_evaluation}}
	\label{fig:ate}
\end{figure}

The evaluation of a single state:

($\angle(\cdot)$ means converting to angle, $log(\cdot)$ returns the canonical coordinates in Lie group)

$$ orientation\ error = \Delta{R_i} =  \angle(\norm[\big]{log(R_{i}\hat{R_i^\prime}^{-1})}) $$
$$ position\ error = \Delta{\mathbf{p}_i} = \norm[\big]{\mathbf{p}_{i} - R_{i}\hat{R_i^\prime}^{-1}\hat{\mathbf{p}}_i^\prime} $$

The root mean square error (RMSE) is then used to quantify the quality of the whole trajectory.
$$ ATE_{rot} = (\frac{1}{N}\sum_{i=o}^{N-1}\Delta{R_i}^2)^{\frac{1}{2}} $$
$$ ATE_{pos} = (\frac{1}{N}\sum_{i=o}^{N-1}\Delta{\mathbf{p}_i}^2)^{\frac{1}{2}} $$

\section{Ground Truth}

I will use the result from \gls{sfm} as the 'ground truth' since \gls{sfm} is currently the most accurate visual reconstruction method. Moreover, the reconstruction area is large to set up a pose tracking system. Thus I used COLMAP to reconstruct all the camera poses with Sift descriptors. The results from COLMAP are used as my 'ground truth' for evaluations of all of the following experiments.

\section{Evaluations}

\subsection{Repeating Pattern, and Rolling Shutter}
Repeating pattern and rolling shutter effect are two common problems that cause mismatches during 2D-2D feature matching and 2D feature to 3D landmark point association which can further influence pose estimation. Because I experimented with a rolling shutter camera in an environment full of repeating patterns (\ref{fig:Klaus Atrium}), therefore I do not specifically design experiments to demonstrate that my pipeline can overcome these two problems. Instead, if I can perform one successful trajectory estimation within my selected environment, a close loop localization result, I can prove that my pipeline overcomes the repeating pattern and rolling shutter problem.

\begin{figure}[ht]
\centering
\subfloat[Handrails]{
\begin{minipage}[t]{0.33\linewidth}
\centering
\includegraphics[width=\textwidth]{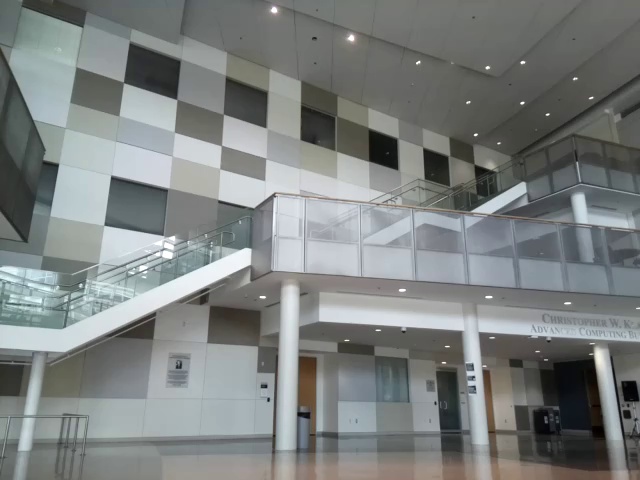}
%\caption{fig1}
\end{minipage}%
}%
\subfloat[Pillars]{
\begin{minipage}[t]{0.33\linewidth}
\centering
\includegraphics[width=\textwidth]{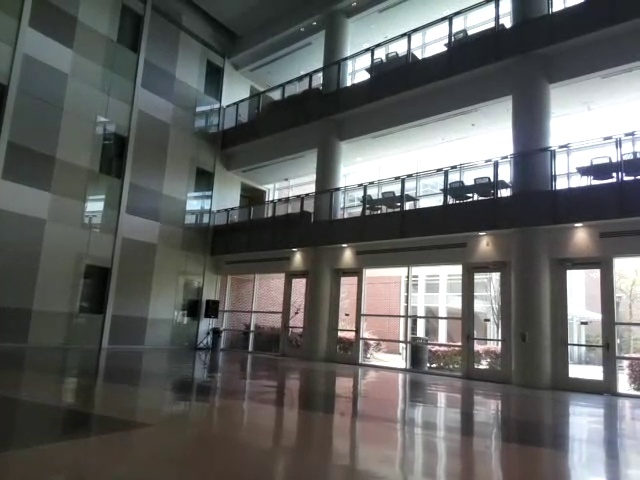}
%\caption{fig2}
\end{minipage}%
}%
\subfloat[Wall Patterns]{
\begin{minipage}[t]{0.33\linewidth}
\centering
\includegraphics[width=\textwidth]{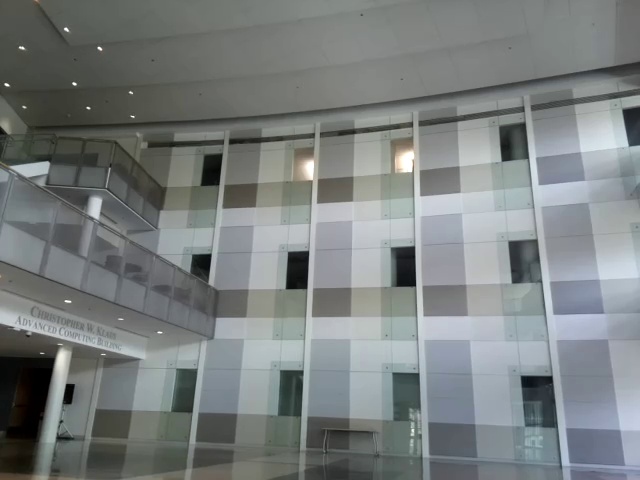}
%\caption{fig2}
\end{minipage}
}%
\centering
\caption{Klaus Atrium}
\label{fig:Klaus Atrium}
\end{figure}

\subsection{Traditional Descriptor vs Learning Descriptor}
I designed my first experiment to compare the effect of the descriptor on my pipeline. To analyze the results between traditional descriptor and learning-based descriptor, I selected Root Sift as the traditional descriptor representative due to its popularity and outstanding performance among visual based localization. Because my pipeline runs with SuperPoint by default, hence the comparison is between Root Sift and SuperPoint.

\subsubsection{Experiment Setup}
I collected a training dataset and a test dataset. Both datasets are collected in a similar trajectory and under same illumination. The datasets are collected by holding the perception system by hand. The training dataset, which is used to reconstruct the sparse point cloud model, contains 137 images. I used the training dataset to reconstructed two sparse point cloud models, one with Sift feature descriptor and one with SuperPoint feature descriptors (\ref{fig:Reconstruction Result}). I used the Sift point cloud to perform localization with Root-Sift and used the SuperPoint point cloud to perform localization with SuperPoint.

\subsubsection{Result}

\begin{figure}[ht]
\centering
\subfloat[Root Sift]{
\begin{minipage}[t]{0.5\linewidth}
\centering
\includegraphics[width=\textwidth]{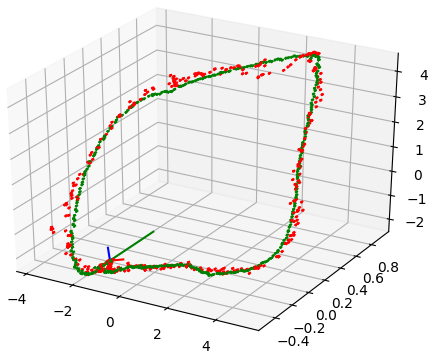}
%\caption{fig1}
\end{minipage}%
}%
\subfloat[SuperPoint]{
\begin{minipage}[t]{0.5\linewidth}
\centering
\includegraphics[width=\textwidth]{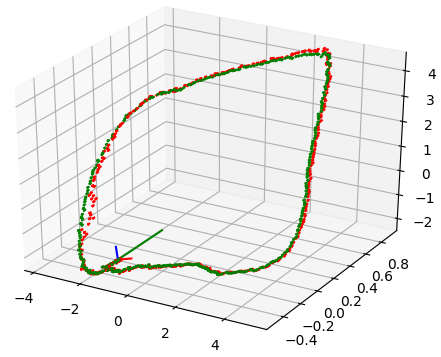}
%\caption{fig2}
\end{minipage}%
}%

\centering
\caption{Root Sift and SuperPoint}{Red is the aligned estimate trajectory. Green is the ground truth trajectory. The aligned estimate trajectory is downsampled from the actual trajectory and contains 387 images.}
\label{fig:Root Sift and SuperPoint}
\end{figure}

\begin{figure}[ht]
    \centering
	\includegraphics[width=\textwidth]{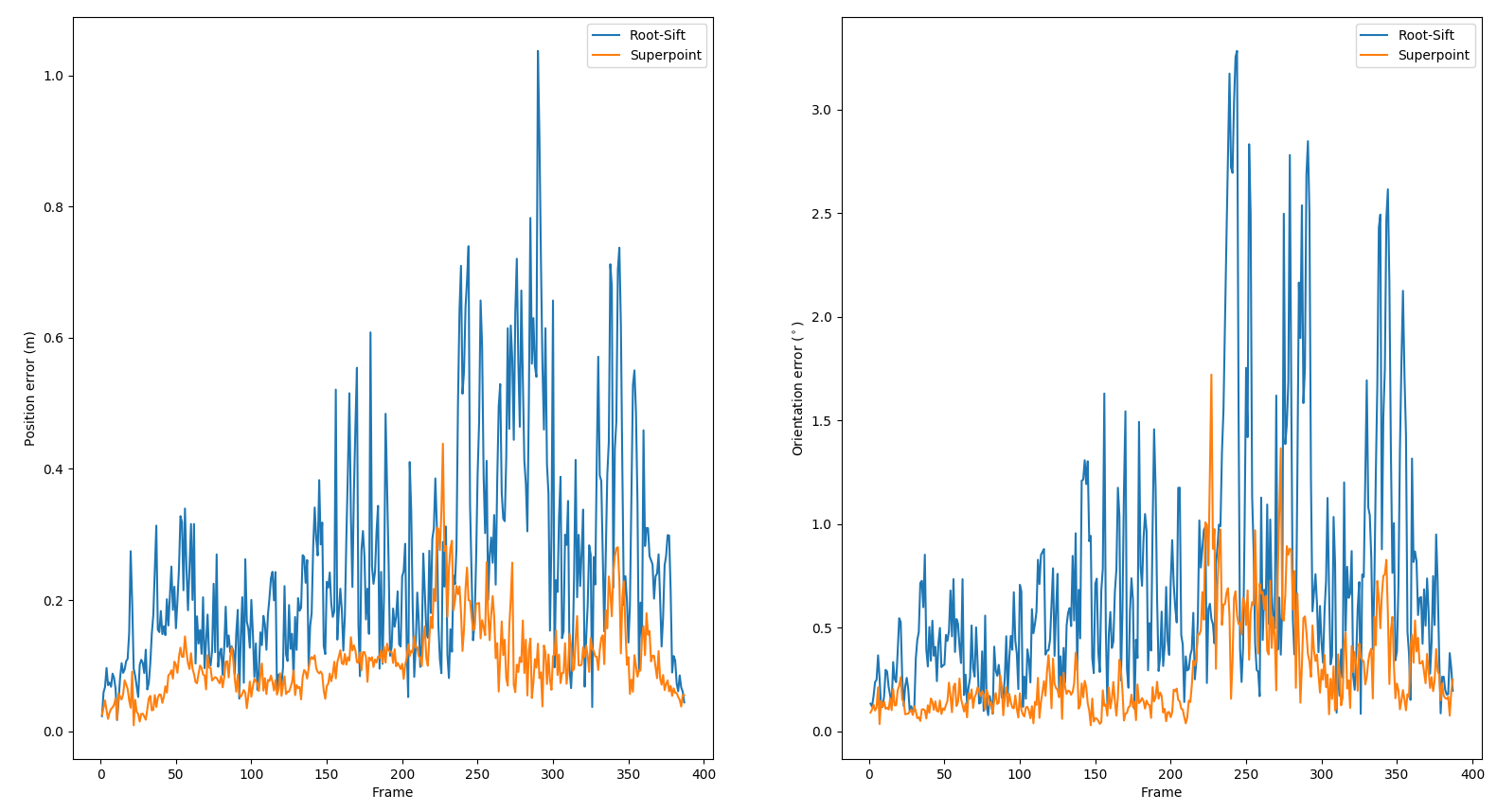}
	\caption[Descriptor error]{SuperPoint and Root Sift trajectory errors}
	\label{fig:descriptor error}
\end{figure}

\begin{table}[h!]
\centering
 \captionof{table}{ATE results of SuperPoint and Root-Sift}\label{tbl:1}
 \begin{tabular}{|c| c| c|} 
 \hline
  & $ATE_{pos} (m)$ & $ATE_{rot} (^\circ)$\\ [0.5ex] 
 \hline
 SuperPoint & 0.123 & 0.369 \\ 
 \hline
 Root-Sift & 0.305 & 0.955 \\
 \hline
 \end{tabular}

\end{table}

\subsubsection{Discussion}
From the result, we can observe that both the Root Sift trajectory result and SuperPoint trajectory result are closed to the ground truth trajectory result. But SuperPoint demonstrates higher localization accuracy than Root Sift.

\subsection{Illumination}
\subsubsection{Experiment Setup}
I collected one training dataset and four test datasets. All datasets are collected under similar trajectories but different illumination conditions. I used the training data to build the 3D model and executed the localization pipeline on all test datasets with SuperPoint descriptor.

\begin{figure}[ht]
    \centering
	\includegraphics[width=\textwidth]{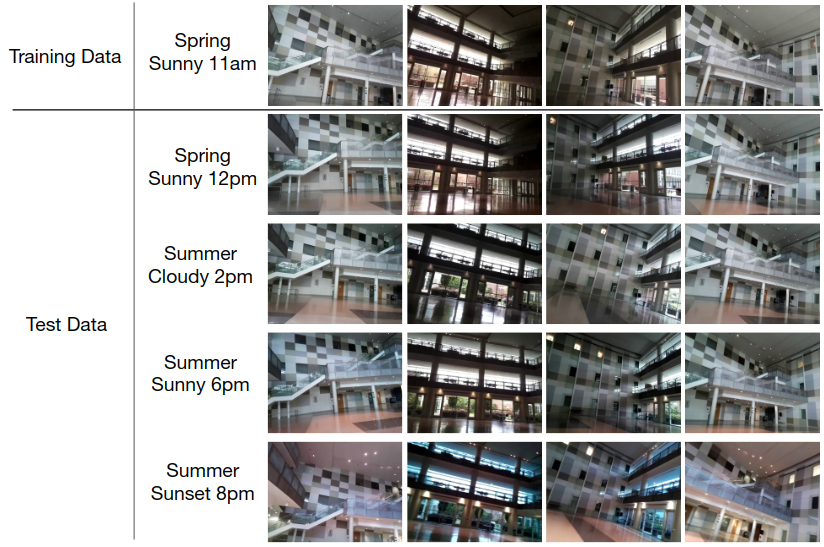}
	\caption[Datasets under different illuminations]{Datasets under different illuminations}
	\label{fig:illuminations}
\end{figure}

\subsubsection{Result}

\begin{figure}[ht]
\centering
\subfloat[Spring Sunny 12pm]{
\begin{minipage}[t]{0.48\linewidth}
\centering
\includegraphics[width=\textwidth]{figures/evaluation/descriptor_SuperPoint.png}
%\caption{fig1}
\end{minipage}%
}%
\subfloat[Summer Cloudy 2pm]{
\begin{minipage}[t]{0.48\linewidth}
\centering
\includegraphics[width=\textwidth]{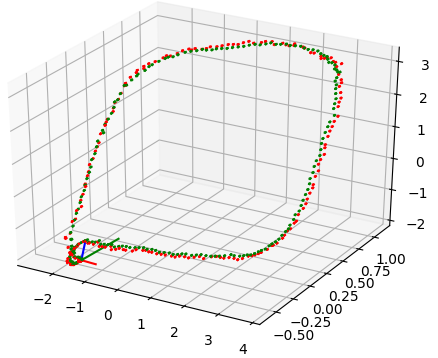}
%\caption{fig1}
\end{minipage}%
}\\
\subfloat[Summer Sunny 6pm]{
\begin{minipage}[t]{0.48\linewidth}
\centering
\includegraphics[width=\textwidth]{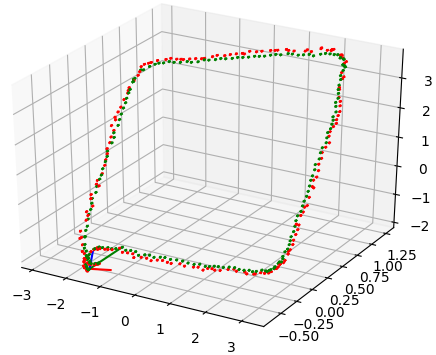}
%\caption{fig1}
\end{minipage}%
}%
\subfloat[Summer Sunset 8pm]{
\begin{minipage}[t]{0.48\linewidth}
\centering
\includegraphics[width=\textwidth]{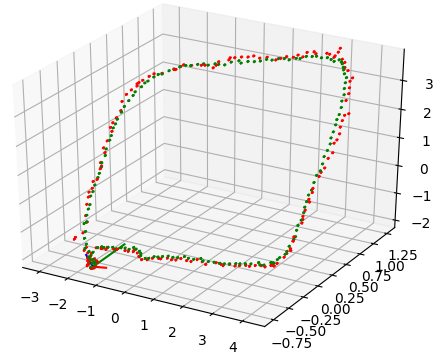}
%\caption{fig1}
\end{minipage}%
}%
\centering
\caption{Localization results under different illuminations}{Red is the aligned estimate trajectory. Green is the ground truth trajectory. The lengths of different test datasets are different.}
\label{fig:illumination}
\end{figure}

\begin{table}[H]
\centering
\captionof{table}{ATE results of different illuminations}\label{tbl:2}
 \begin{tabular}{|c|c|c|} 
 \hline
  \multirow{2}{*}{Datasets} & \multicolumn{2}{|c|}{SuperPoint Descriptor} \\ \cline{2-3}
  & $ATE_{pos} (m)$ & $ATE_{rot} (^\circ)$\\ [0.5ex] 
 \hline
 Spring Sunny 12pm & 0.123 & 0.369 \\ 
 \hline
 Summer Cloudy 2pm & 0.255 & 0.669 \\
 \hline
 Summer Sunny 6pm & 0.31 & 0.622 \\
 \hline
  Summer Sunset 8pm & 0.298 & 0.714 \\
 \hline
 \end{tabular}
\end{table}

\subsubsection{Discussion}
The result from \ref{tbl:2} and \ref{fig:illuminations} indicates that my pipeline can run on datasets that are collected under illumination different from the training data. The Spring Sunny 12 pm dataset is the most accurate because the trajectory and illumination are similar to the training dataset. The other three datasets contain more degenerated frames caused by the motion blur effect. These degenerated frames increased the error of the final result.

\subsection{Large field of view changes and 3D Model Density}
\subsubsection{Experiment Setup}
To test how viewpoint changes will affect the localization result, I collected the training data in a large circle with all the camera facing inside the Atrium. Then, I collected the test data in a small circle with all the camera facing forward. I later collected additional images that can be added to the training data to reconstruct 3D models of two different densities.

\subsubsection{Result}

\begin{figure}[ht]
\centering
\subfloat[Sparse Map]{
\begin{minipage}[t]{0.48\linewidth}
\centering
\includegraphics[width=\textwidth]{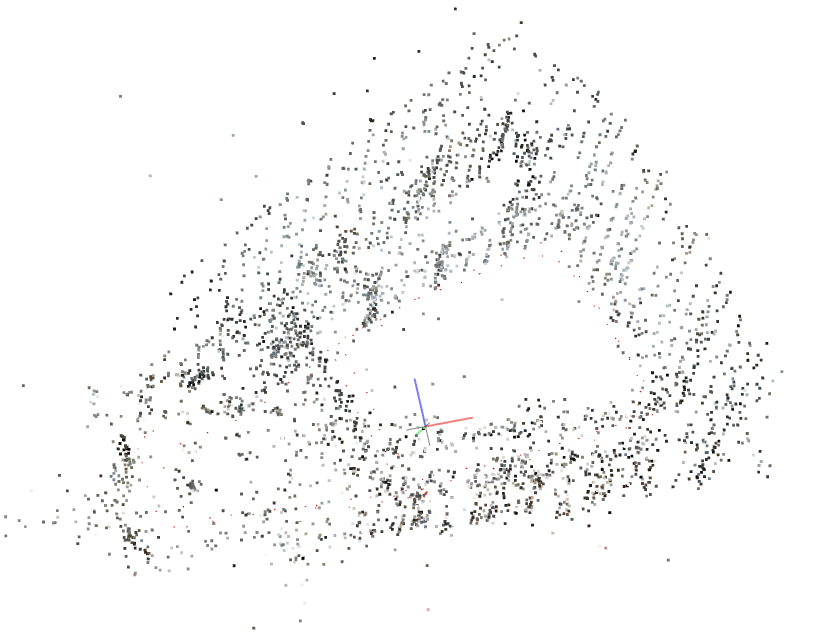}
%\caption{fig1}
\end{minipage}%
}%
\subfloat[Dense Map]{
\begin{minipage}[t]{0.48\linewidth}
\centering
\includegraphics[width=\textwidth]{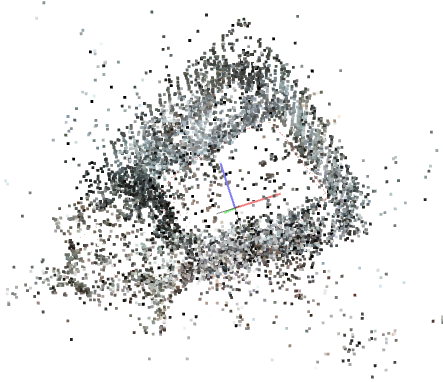}
%\caption{fig1}
\end{minipage}%
}
\centering
\caption{Map with different density}{The sparse map is reconstructed with 126 images and contains 4077 points. The dense map is reconstructed with 513 images and contains 10507 points.}
\label{fig:density}
\end{figure}

\begin{figure}[ht]
\centering
\subfloat[Sparse Map trajectory]{
\begin{minipage}[t]{0.48\linewidth}
\centering
\includegraphics[width=\textwidth]{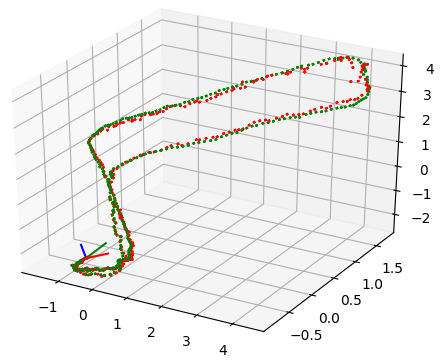}
%\caption{fig1}
\end{minipage}%
}%
\subfloat[Dense Map trajectory]{
\begin{minipage}[t]{0.48\linewidth}
\centering
\includegraphics[width=\textwidth]{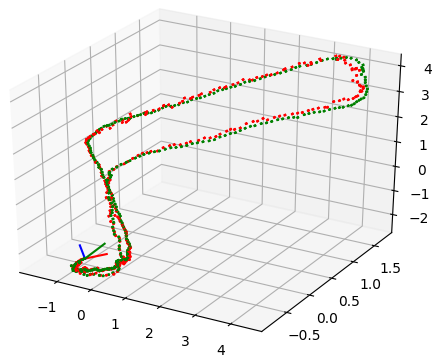}
%\caption{fig1}
\end{minipage}%
}
\centering
\caption{Trajectory with different density maps}{Red is the aligned estimate trajectory. Green is the ground truth trajectory.}
\label{fig:density trajectory}
\end{figure}

\begin{table}[H]
\centering
\captionof{table}{ATE results of different illuminations}\label{tbl:3}
 \begin{tabular}{|c|c|c|} 
 \hline
  \multirow{2}{*} & Sparse Map & Dense Map\\ [0.5ex] 
 \hline
 $ATE_{pos} (m)$ & 0.311 & 0.268 \\ 
 \hline
 $ATE_{rot} (^\circ)$ & 0.701 & 0.567 \\
 \hline
average SuperPoint extraction time $(s)$ & 0.0831079 & 0.0711602 \\
 \hline
  average landmark projection time $(s)$ & 0.000665298 & 0.00112228 \\
   \hline
  average success projected landmarks & 1089.8 & 2614.35  \\
 \hline
   average 2D-3D data association time $(s)$ & 0.159091 & 0.1665 \\
  \hline
    average pose estimation time $(s)$ & 0.0126785 & 0.0139296 \\
 \hline
    average total time $(s)$ & 0.260967 & 0.256019 \\
 \hline
 \end{tabular}
\end{table}

\subsubsection{Discussion}
As seen in \ref{fig:density trajectory}, the pipeline can estimate the trajectory even when the test data viewpoint is much different from the training dataset. This is because SuperPoint descriptors are invariant to viewpoint changes. Thus 2d features and 3d landmarks can still be matched together even when their associated descriptors are generated under different viewpoints.

As for how the map density will affect the localization result, from \ref{tbl:3} we can see that localization under dense map not only results in higher accuracy but also surprisingly results in lower computation time than localization under sparse map. This is because the main factor of the computation time is SuperPoint extraction time and 2D-3D data association time. The density of the map does not affect the feature extraction step. It also has less effect on 2D-3D data association time because in my pipeline within the 2D-3D data association step, I iterative through all SuperPoint features instead of projected landmarks.

\subsection{Motion Model}

\subsubsection{Experiment Setup}
I hope to compare how different motion models will affect the localization result. Hence, apart from the static motion model, which is the default motion model used in my pipeline, I also applied a constant speed motion model in the experiment. I used the same dataset from the 3D model density experiment to complete this experiment.

Static motion model:
\begin{equation}
    {}_wT_{k+1} = {}_wT_{k} 
\end{equation}

Constant speed motion model:
\begin{equation}
\begin{split}
    {}_{k+1}T_{k} = {}_kT_{k-1}\\
    {}_wT_{k+1} = {}_wT_{k} * {}_kT_{k-1}
\end{split}
\end{equation}

where ${}_wT_{k}$ is the 6 \gls{dof} pose in the world coordinate system at $k$ time step. ${}_kT_{k-1}$ is the transformation from $k-1$ time step pose to the $k$ time step pose.

\subsubsection{Result}
\begin{table}[h]
\centering
\captionof{table}{Map Density and Motion Model}\label{tbl:4}
 \begin{tabular}{|c| c| c|} 
 \hline
  & $ATE_{pos} (m)$ & $ATE_{rot} (^\circ)$\\ [0.5ex] 
 \hline
 Sparse Map Static Speed & 0.704 & 0.311\\ 
 \hline
 Sparse Map Constant Speed & 0.640 & 0.311 \\
 \hline
 Dense Map Static Speed & 0.567 & 0.268 \\
 \hline
 Dense Map Constant Speed & 0.501 & 0.265 \\
 \hline
 \end{tabular}
\end{table}

\begin{figure}[ht]
\centering
\subfloat[Sparse map static motion model trajectory]{
\begin{minipage}[t]{0.48\linewidth}
\centering
\includegraphics[width=\textwidth]{figures/evaluation/sparse_traj.png}
%\caption{fig1}
\end{minipage}%
}%
\subfloat[Sparse map constant speed motion model trajectory]{
\begin{minipage}[t]{0.48\linewidth}
\centering
\includegraphics[width=\textwidth]{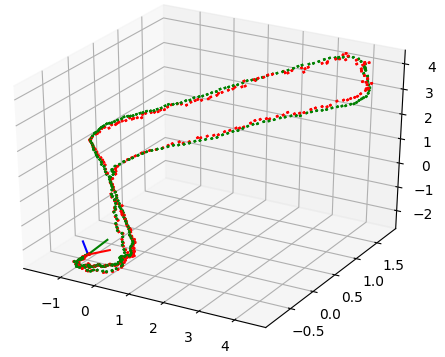}
%\caption{fig1}
\end{minipage}%
}\\
\subfloat[Dense map static speed motion model trajectory]{
\begin{minipage}[t]{0.48\linewidth}
\centering
\includegraphics[width=\textwidth]{figures/evaluation/dense_traj}
%\caption{fig1}
\end{minipage}%
}
\subfloat[Dense map constant speed motion model trajectory]{
\begin{minipage}[t]{0.48\linewidth}
\centering
\includegraphics[width=\textwidth]{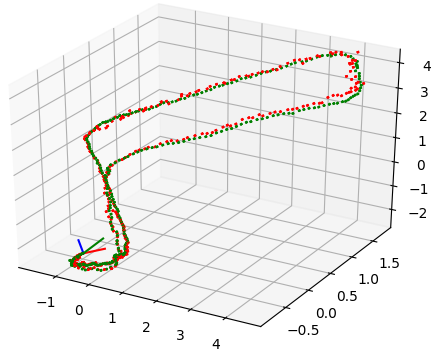}
%\caption{fig1}
\end{minipage}%
}
\centering
\caption{Trajectory with different motion model}{Red is the aligned estimate trajectory. Green is the ground truth trajectory.}
\label{fig:motion model trajectory}
\end{figure}
\subsubsection{Discussion}
As shown in \ref{tbl:4} and \ref{fig:motion model trajectory}, a better motion model result in a more accurate trajectory because it provides better pose prior and 2D-3D association during the pose estimation. The best combination for my pipeline is a dense map combine with a constant speed motion model.

\subsection{Fast-Motion Dataset}
\subsubsection{Experiment Setup}
In all previous experiments, all datasets are collected at a slow-motion because the localization system is designed for a slow-motion blimp that moves around at the speed of approximately $1 m/s$ for translation and $10 ^\circ/s$ for rotation. 

In this experiment, I hope to test how a fast-motion dataset input will affect the localization result. Based on the previous experiments, I know that a density map with a constant speed model is the best combination for my pipeline. Thus, I collected a training dataset with 515 images to reconstruct a dense map. The dense map contains 10,591 points. The test dataset is collected at the speed of around $4, \text{m/s}$ for translation and $45^\circ/\text{s}$ for rotation. Because of the fast motion, the dataset contains a lot of frames with strong motion blur and rolling shutter effects.

\begin{figure}[ht]
\centering
\subfloat{
\begin{minipage}[t]{0.33\linewidth}
\centering
\includegraphics[width=\textwidth]{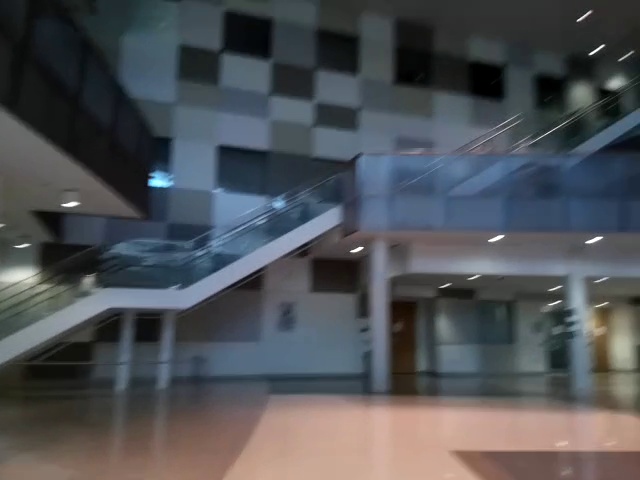}
%\caption{fig1}
\end{minipage}%
}%
\subfloat{
\begin{minipage}[t]{0.33\linewidth}
\centering
\includegraphics[width=\textwidth]{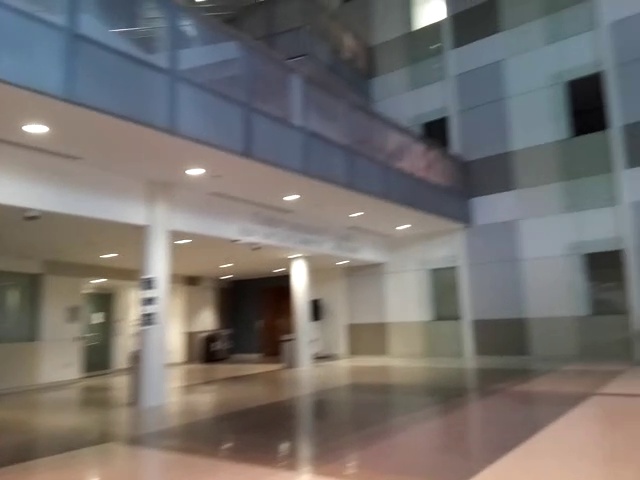}
%\caption{fig1}
\end{minipage}%
}
\subfloat{
\begin{minipage}[t]{0.33\linewidth}
\centering
\includegraphics[width=\textwidth]{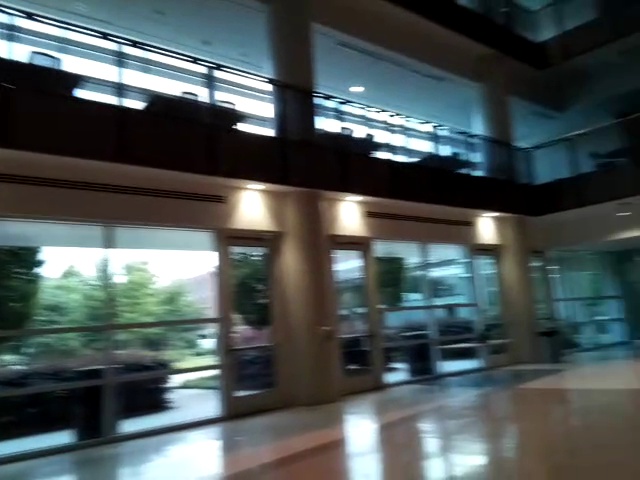}
%\caption{fig1}
\end{minipage}%
}
\centering
\caption{Motion Blur Frames}
\label{fig:motion_blur}
\end{figure}

\subsubsection{Result}
\begin{figure}[ht]
\centering
\subfloat{
\begin{minipage}[t]{\linewidth}
\centering
\includegraphics[width=0.7\textwidth]{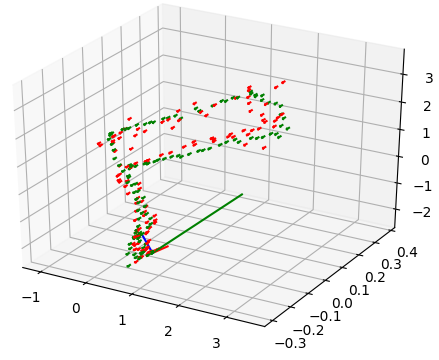}
%\caption{fig1}
\end{minipage}%
}%
\centering
\caption{Fast Motion Dataset Result}
\label{fig:fast motion}
\end{figure}

\begin{table}[h!]
\centering
\captionof{table}{ATE results of fast motion dataset}\label{tbl:5}
 \begin{tabular}{|c| c|} 
 \hline
  $ATE_{pos} (m)$ & $ATE_{rot} (^\circ)$\\ 
 \hline
 1.178 & 0.379 \\ 
 \hline
 \end{tabular}
\end{table}

\subsubsection{Discussion}
From the result, we can observe that the pipeline can only roughly estimate the trajectory of the camera with a high error. This result shows that my pipeline is not robust to fast-motion camera movement, because the motion-blur effect that normally exists for several frames causes translation and rotation drift during pose estimation.

    \chapter{Conclusion}

Vision localization based on a prebuilt map is becoming increasingly popular recently. In this thesis, an indoor incremental localization pipeline has been implemented based on SuperPoint descriptor. The pipeline utilizes a motion model to associate 3D landmarks with 2D features. In this way, I developed an incremental localization solution based on conventional global vision-based localization structures. The pipeline can continuously estimate camera poses based on a sparse map and an initial pose. Several experiments have been conducted on the pipeline. The results indicate that the pipeline is robust to illumination and viewpoint changes. And the pipeline is not affected by repeating patterns and rolling shutter effects. By comparing localization results on different descriptors, it is clear that SuperPoint based localization is more accurate than and Root Sift based localization. Finally, my raspberry pi based perception system can be easily integrated into blimp or other miniature low-cost robots.
    \chapter{Future Work}

In this thesis, I have presented some results for incremental localization in an indoor environment. The developed pipeline has a lot of potential for improvements as it is only a first implementation where the amount of optimization has been minimal. For example, this pipeline requires an initial pose to begin trajectory estimation. In this case, pose estimation failure will terminate the trajectory estimation process. To order to solve this problem, I can integrate a global localization module into the system. 

Moreover, I believe each pose estimation factor graph can be added to a large factor graph constructed by the points and poses from the \gls{sfm} process. In this way, the occlusion problem can be solved by considering each observation's observability during the pose estimation. 

In addition, the current implementation sets the minimal descriptor distance within the 2D-3D association as a fixed value. I believe this value can be studied during the localization process. By doing so, the 2D-3D association process would contain fewer outliers and result in more accurate pose estimation.

As applications for incremental localization usually requires real-time performance, the next step would be to create a pipeline capable of real-time localization and record benchmarks for comparison to other localization methods. To increase computation speed, I can convert loop-based modules written in python into C++ code or use a more advanced desktop than a laptop. I can further study the focus of features in the image to reduce computation on the 2D-3D association. 

Finally, the current experiments are not tested on the blimp because I have not developed the blimp control system. In the future, operations can be conducted by attaching the hardware system on a blimp. If the system runs on the blimp, I hypothesis that one of the main differences is the input sequence will contain more blurry effects caused by the unstable movement of the blimp. Luckily this problem can be eliminated if we know the motion model of the blimp.
    % \makeBibliography
\clearpage
\bookmarksetupnext{level=part}
\phantomsection
\addcontentsline{toc}{chapter}{References}
\renewcommand{\bibname}{References}
\bibliography{references}

\end{thesisbody}

\end{document}